\documentclass[10pt,conference,a4paper]{IEEEtran}
\usepackage{times}
\usepackage{graphicx}
\usepackage{subfigure}
\usepackage{amsmath}
\usepackage{tabularx}
\usepackage{amssymb}
\usepackage{mathrsfs}
\usepackage{algorithm}
\usepackage{algorithmic}
\usepackage{multirow}
\usepackage{bm}
\usepackage{url}
\title{Designing A Composite Dictionary Adaptively From Joint Examples}
\author{%
{Zhangyang Wang\small $~^\dag$}, Yingzhen Yang{\small $~^\dag$}, Jianchao Yang{\small $~^\ddag$}, Thomas Huang{\small $~^\dag$} 
{}
\vspace{1.6mm}\\
\fontsize{10}{10}\selectfont\itshape
\dag Beckman Institute, University of Illinois at Urbana-Champaign, Urbana, IL 61801, USA\\ 
\ddag Snapchat Inc, Venice, CA 90291, USA\\
\fontsize{9}{9}\selectfont\ttfamily\upshape
\vspace{1.2mm}\\
\fontsize{10}{10}\selectfont\rmfamily\itshape
\{zwang119, yyang58\}@illinois.edu, jianchao.yang@snapchat.com, t-huang1@illinois.edu
}
\begin{document}
\maketitle

\begin{figure}[b]
\parbox{\hsize}{\em
IEEE VCIP'15, Dec. 13 - Dec. 16, 2015, Singapore

000-0-0000-0000-0/00/\$31.00 \ \copyright 2015 IEEE.
}\end{figure}

\begin{abstract}
We study the complementary behaviors of external and internal examples in image restoration, and are motivated to formulate a composite dictionary design framework. The composite dictionary consists of the global part learned from external examples, and the sample-specific part learned from internal examples. The dictionary atoms in both parts are further adaptively weighted to emphasize their model statistics. Experiments demonstrate that the joint utilization of external and internal examples leads to substantial improvements, with successful applications in image denoising and super resolution. 
\\[1\baselineskip]
\end{abstract}

 
%

\section{Introduction}
\label{sec:intro}
Sparse coding (SC) by representing a signal as a sparse linear combination of representation bases (dictionary atoms) has been widely applied \cite{elad2006image}. The dictionary, which should both faithfully represent the signal and effectively extract task-specific features, plays an important role \cite{rubinstein2010dictionaries}. For image restoration, classical methods either rely on a large external set of image examples \cite{yang2010image}, or find self-similar examples from the input  \cite{Fattal2010}. With much progress being made, it is recently recognized that external and internal examples each suffer from certain drawbacks, but their properties may be complementary \cite{wang2014joint, wang2015learning}. 

We believe the joint utilization of external and internal examples in dictionary design is crucial for further improving image restoration. We thus formulate a new composite dictionary design framework for image restoration tasks. Successful applications in image denoising and super resolution (SR) demonstrate its effectiveness. 



\section{Related Work}
\label{sec:work}
The problem to be investigated resembles to a general problem in SC-based classification: how to adaptively build the relationship between dictionary atoms and class labels? Based on predefined relationships, current supervised dictionary learning (DL) methods are categorized into either learning a shared dictionary by all classes, which may be compact but not sufficiently discriminative \cite{mairal2012task}; or a class-specific dictionary with the opposite properties \cite{mairal2008discriminative}. In \cite{yang2014latent}, the authors jointly learned a composite dictionary combing class-specific and shared dictionary atoms, with a latent matrix indicating the relationship between dictionary atoms and labels.

In analogy to the classification case, reconstruction-purpose dictionaries have been built from either external or internal examples. External exampled-based methods are known for their capabilities to produce plausible image appearances. However, there is no guarantee that an arbitrary input patch can be well matched or represented by a pre-fixed external set. When there is rarely any match for the input, external examples are prone to introduce either noise or oversmoothness  \cite{yang2011exploiting}.  Meanwhile, the self similarity property supplies internal examples that are highly relevant to the input, but only of a limited number. Due to the insufficiency of internal examples, their mismatches often result in severe visual artifacts \cite{chatterjee2010denoising}. 

The joint utilization of both external and self examples has been first studied for image denoising \cite{zontak2011internal}. Mosseri et. al. \cite{mosseri2013combining} proposed that image patches had different preferences towards either external or self examples for denoising. Such a preference is in essence the tradeoff between noise-fitting versus signal-fitting. In \cite{wang2014joint, wang2015learning}, a joint super-resolution (SR) models was proposed to adaptively combine the advantages of both external and self example-based loss functions. \cite{wang2015self} further investigated the utilization of self-similarity into deep learning-based SR. However, none of the prior work makes much progress towards a unified dictionary design framework.

\section{Technical Approach}
\subsection{Overview}
The \textit{composite dictionary} consists of the \textit{global dictionary} part learned from external examples, and the \textit{sample-specific dictionary} part learned from internal examples. The atoms in both parts are further weighted to exploit the different model statistics. Given input signal $\mathbf{x} \in R^{p \times 1}$, the formulation can be mathematically presented as: 
\begin{equation}
\begin{array}{l}\label{e00}
\underset{\mathbf{a},  \omega_G,  \omega_S}{\min} \lambda_E \sum_i ||\mathbf{a^G}_i||_1 + \lambda_I \sum_j ||\mathbf{a^S}_j||_1 +||\mathbf{x} - 
\\ \sum_i \mathbf{d^G_i} M_G(\mathbf{d^G_i, x}, \omega_G) \mathbf{a^G_i}  - \sum_j \mathbf{d^S_j} M_I(\mathbf{d^S_j, x}, \omega_S) \mathbf{a^S_j} ||_F^2 
\end{array}
\end{equation}
Here $\mathbf{d^G}_i \in R^{p \times 1}, i=1,2,..., M$ denotes the dictionary atoms pre-learned from external examples, and $\mathbf{d^S}_j \in R^{p \times 1}, j=1,2,..., N$ the atoms pre-learned from internal examples. We define $\mathbf{D^G}=\{\mathbf{d^G}1,..., \mathbf{d^G}_M\}$ as the \textit{global base dictionary} and $\mathbf{D^S}=\{\mathbf{d^S}_1,..., \mathbf{d^S}_N\}$ as the \textit{sample-specific base dictionary}. $\mathbf{a} \in R^{(M+N) \times 1}$ denotes sparse codes of $\mathbf{x}$, consisting of $\mathbf{a^G}_i, i=1,2,..., M$ and $\mathbf{a^S}_j, j=1,2,..., N$, corresponding to $\mathbf{D^G}$ and $\mathbf{D^S}$, respectively. $\lambda_E$ and $\lambda_I$ are constants. Note that both the first two terms, and the last two summations in the third term of (\ref{e00}), can be each combined together just like in conventional SC. We prefer writing them separately in a purpose to highlight two different dictionary parts.  $M_G$ and $M_I$ denote some similarity-based weights between the dictionary atoms and the input, parametrized by $\omega_G$ and  $\omega_S$, respectively. 

Our solution to (\ref{e00}) takes three steps: 
\begin{itemize}
\item Obtain $\mathbf{D^G}$ and $\mathbf{D^S}$ prior to solving (\ref{e00}).
\item Choose the desired forms of $M_G$ and $M_I$.
\item Solve (\ref{e00}) using a coordinate-descent algorithm. 
\end{itemize}
Note the strategy is to first fix base dictionaries, then adapting them to the input by learning weights, which is close to \cite{rubinstein2010double}. Both lead to efficient and flexible dictionary representations. Yet rather than simply enforcing sparsity constraints on the weights \cite{rubinstein2010double}, we aim to build a more adaptive relationship between the input and the atoms, based on the complementary example statistics. Additionally, while the base in \cite{rubinstein2010double} is simply a DCT dictionary, our $\mathbf{D^G}$ and $\mathbf{D^S}$ are specifically crafted from external and internal examples separately. 


\begin{algorithm}[ht]
\caption{Coordinate descent algorithm for solving (\ref{e00})}
\begin{algorithmic}[1]
\REQUIRE $\mathbf{D^G}$ and $\mathbf{D^S}$; $\lambda_E$ and $\lambda_I$; \ ITER; $\beta$.

\STATE FOR t=1 to ITER DO

\STATE Fix $\mathbf{\Omega_G}$ and $\mathbf{\Omega_S}$, solve (\ref{e00}) over $\mathbf{A}$ using the feature-sign algorithm \cite{lee2006efficient}. 

\STATE Fix $\mathbf{\Omega_G}$ and $\mathbf{A}$, solve $\mathbf{\Omega_S}$ by taking gradient descent over $\mathbf{F_S}$, with step size $\beta$.

\STATE Fix $\mathbf{\Omega_S}$ and $\mathbf{A}$, solve $\mathbf{\Omega_G}$ by taking gradient descent over $\mathbf{F_G}$, with step size $\beta$.

\STATE END FOR

\ENSURE $\mathbf{A}$, $\mathbf{\Omega_G}$ and $\mathbf{\Omega_S}$
\end{algorithmic}
\end{algorithm}


\subsection{Algorithm}
First of all, we learn $\mathbf{D^G}$ and $\mathbf{D^S}$ from the sets of external and internal examples respectively (e.g., by K-SVD \cite{elad2006image}). Before moving on to learn functional forms of $M_G$ and $M_I$, it is interesting to examine whether a static, but well-defined weight could help. Without loss of generality, we assume both $M_G$ and $M_I$ have a value range between [0,1].  It is obvious that, when $M_G$ (or $M_I$) becomes larger, i.e., the current atom is highly correlated to the input, the atom will be more favored by the $\ell_1$ penalty. We first define $M_G$ and $M_I$ both in the form of  radial basis function (RBF) kernels:
\begin{equation}
\begin{array}{l}\label{plain}
M_G(\mathbf{d^G_i, x}, \omega_G) = \exp(- \omega_G ||\mathbf{d^G_i} - \mathbf{x}||^2)\\
M_I(\mathbf{d^S_j, x}, \omega_S) = \exp(- \omega_S ||\mathbf{d^S_j} - \mathbf{x}||^2)
\end{array}
\end{equation}
where $\omega_G$ and $\omega_S$ are both fixed constants, but of different values. As discussed above, it is often more likely to find ``highly similar" examples internally than from external examples. On the other hand, the external set can usually provide more abundant ``reasonably similar" (not necessarily highly though) examples. We thus desire the value of $M_I$ to decrease more quickly than $M_G$, which means $\omega_S$ is supposed to be chosen larger than $\omega_G$. Note that when $\omega_G$ and $\omega_S$ become fixed, (\ref{e00}) becomes a plain sparse decomposition problem, that can be solved efficiently \cite{lee2006efficient}.


Inspired by (\ref{plain}), it is straightforward to construct parameterized  $M_G$ and $M_I$ in the form of Mahalanobis kernel \cite{kamada2006support}: 
\begin{equation}
\begin{array}{l}\label{adaptive}
M_G(\mathbf{d^G_i, x}, \mathbf{\Omega_G} ) = \exp(- (\mathbf{d^G_i} - \mathbf{x})^T \mathbf{\Omega_G} (\mathbf{d^G_i} - \mathbf{x}))\\
M_I(\mathbf{d^S_j, x}, \mathbf{\Omega_S} ) = \exp(-(\mathbf{d^S_j} - \mathbf{x})^T  \mathbf{\Omega_S} (\mathbf{d^S_j} - \mathbf{x}))
\end{array}
\end{equation}
Note $\mathbf{\Omega_G}$ and $\mathbf{\Omega_S}$ are both semi-definite real matrices. However, learning $\mathbf{\Omega_G}$ or $\mathbf{\Omega_S}$ directly requires enforcing a positive semi-definite constraint during optimization, which is expensive. A cheaper and well-known solution is to decompose: $\mathbf{\Omega_G} = \mathbf{F_G}^T \mathbf{F_G}$, $\mathbf{\Omega_S}$ = $\mathbf{F_S}^T \mathbf{F_S}$. Since $\mathbf{F_G}$ is an unconstrained real matrix, we can now cast the metric learning as an unconstrained matrix optimization problem. 

Concluding all above, we solve (\ref{e00}) by a coordinate descent algorithm, as detailed in Algorithm 1.


\section{Experiments}

In our experiments, $\mathbf{x}$ is by default a $5 \times 5$ image patch, columnized to be a $25 \times 1$ vector; the external/internal examples and resulting dictionary atoms share the same dimension. We use the natural patches cropped from the Berkley Segmentation Dataset (BSD) as the collection of external examples. The internal example candidates are cropped from the input image with a spatial overlap of 1, to ensure a reasonably sufficient amount. For image SR, the methodology is similar but works on example pairs. We set $M=128$, and $N=32$ as the default dictionary sizes of $\mathbf{D^G}$ and $\mathbf{D^S}$.  

In Algorithm 1,  $\mathbf{F_G}$ and $\mathbf{F_S}$ are both initialized to be identity matrices with some random disturbance on the diagonal elements. When handling (\ref{plain}), we fine-tune $\omega_G$ and $\omega_S$ by cross-validation. The solved $\mathbf{A}$ from (\ref{plain}) serves as a proper initialization in (\ref{adaptive}). $\lambda_E$ and $\lambda_I$ vary by applications and will be tuned, but we find it universally applicable to set  $\lambda_I$ around 10 times of  $\lambda_E$. $\beta$ is fixed at 0.9, and $ITER$ is 5 for all. When well initialized, the current MATLAB implementation takes no more than 5 iterations to converge, and each iteration consumes 10-15 minutes for a 256 $\times$ 256 image.

\subsection{KNN versus KSVD: The Power of Weights}
As the choice of base dictionaries can be quite flexible per application requirements, it is interesting to evaluate if we could rely on simple, computationally cheap bases for a comparable performance to the sophisticated ones. 

We construct the ``KNN base dictionaries'':  for $\mathbf{D^G}$, we use a K-NN clustering over the external examples and $M$ cluster centroids are obtained. For $\mathbf{D^S}$, we simply obtain $N$ closest internal examples from the input image. We then compare the following methods in a typical image denoising setting:
\begin{itemize}
\item \textbf{Method I.} Solving (\ref{e00}) using $\mathbf{D^G}$ and $\mathbf{D^S}$
\item \textbf{Method II.} Performing conventional SC over the composite dictionary of $\mathbf{D^G}$ and $\mathbf{D^S}$ (equivalent to let $M_G=M_I=\mathbf{I}$), as a benchmark.
\item \textbf{Method III.}  The K-SVD algorithm \cite{elad2006image} is first applied to either external or internal examples, to obtain the global and sample-specific K-SVD dictionaries, respectively. The two K-SVD dictionaries are concatenated into a composite dictionary, over which SC is performed.
\end{itemize}
Five natural images, \textit{Lena}, \textit{Barbara}, \textit{Boats}, \textit{House} and \textit{Peppers} are used for testing, with gaussian noise of standard deviation $\sigma$ = 10. As in Table I, it is impressive to see that Method I, which relies on re-weighting the simplest KNN base dictionaries in (\ref{e00}), outperforms the canonical KSVD dictionaries. We also see a large average margin of 4dB of Method I over Method II in terms of PSNR, which clearly manifests the benefits of modeling and learning proper weights.

 \begin{table}[h]
 \begin{center}
 \caption{Comparison of PNSRs (dB) used three different methods.}
 \vspace{1em}
 \begin{tabular}{|c|c|c|c|c|c|}
 \hline
 &  \textit{Lena} &  \textit{Barbara} & \textit{Boats} & \textit{House} &  \textit{Peppers}  \\
 \hline
Method I & \textbf{35.57} &  33.98  & \textbf{33.83} & 33.56 & \textbf{34.93} \\
 \hline
Method II & 31.21 &  30.41 & 31.24 &  29.43 & 30.67 \\
 \hline
Method III & 35.36 & \textbf{34.24} & 33.62 &  \textbf{34.76} & 34.32 \\
 \hline
 \end{tabular}
 \end{center}
 \end{table}
 


\subsection{Application I: Image Denoising}

Image denoising is a most classical application scenario for SC and DL. Each image is processed in a patch-wise manner with a spatial overlap of 1. For the best performances, we use the global and sample-specific K-SVD dictionaries, obtained in the above Method III, as our base dictionaries $\mathbf{D^G}$ and $\mathbf{D^S}$. We compare the following methods on the five natural images:
\begin{itemize}
\item \textbf{KSVD G} denotes SC directly performed over the global K-SVD dictionary $\mathbf{D^G}$.
\item \textbf{KSVD S} denotes SC directly performed over the sample -specific K-SVD dictionary $\mathbf{D^S}$.
\item \textbf{KSVD C} denotes SC directly performed over the composite dictionary of $\mathbf{D^G}$ and $\mathbf{D^S}$.
\item \textbf{SC FW} denotes ``SC with fixed weights" by solving (\ref{e00}) over the composite dictionary of $\mathbf{D^G}$ and $\mathbf{D^S}$, with $M_G$ and $M_I$ defined in (\ref{plain}).
\item \textbf{SC LW} denotes ``SC with learned weights" by solving (\ref{e00}) over the composite dictionary of $\mathbf{D^G}$ and $\mathbf{D^S}$, with $M_G$ and $M_I$ defined in (\ref{adaptive}).
\end{itemize}
$\sigma$ varies from 10 to 50, with a stride of 10. For each method, the \textit{average} PNSRs over all five images under various $\sigma$ values are reported in Table II. The proposed SC LW outperforms all else with a large margin of around 2dB, in all cases. 

 \begin{table}[h]
 \begin{center}
 \caption{Comparison of PNSRs (dB) among different denoising methods.}
 \vspace{1em}
 \begin{tabular}{|c|c|c|c|c|c|}
 \hline
 &  $\sigma$ = 10 &  $\sigma$ = 20 & $\sigma$ = 30  & $\sigma$ = 40 &  $\sigma$ = 50  \\
 \hline
KSVD G & 33.57 &  30.18  & 28.83 & 26.43 & 25.32 \\
 \hline
KSVD S & 34.23 &  31.02 & 28.94 &  26.66 & 25.48 \\
 \hline
KSVD C & 34.46 & 32.24 & 29.62 &  26.76 & 25.67 \\
 \hline
SC FW & 34.83 & 33.45 & 30.28 &  26.27 & 25.32 \\
 \hline
SC LW & \textbf{36.27} & \textbf{34.24} &  \textbf{32.83} &  \textbf{28.76} &  \textbf{26.32} \\
 \hline
 \end{tabular}
 \end{center}
 \end{table}
 
The learned weights help SC LW outperform SC FW remarkably. When $\sigma$ goes larger, the performances of SC FW degrade quickly. It can be interpreted that the (static) RBF weights become less reliable in describing the correlations between the noisy input patch and dictionary atoms, especially those from $\mathbf{D^G}$ which are cropped from noise-free images. In contrast, the learned weights show better robustness. Also, it is not a surprise to see that the utilization of joint examples leads to the consistent superiority of KSVD C over either KSVD G or KSVD S.

Interestingly, by comparing KSVD S and KSVD G, we observe that internal examples gain advantages over external ones under small $\sigma$s. For large $\sigma$s, the results of KSVD S deteriorate faster and become worse than KSVD G when $\sigma$ = 50. The performance margin of KSVD C over KSVD G is also reduced when $\sigma$ increases. Such observations imply that when the noise becomes heavy, internal examples are overly corrupted and cannot provide relevant references well. That coincide with the conclusion in \cite{elad2006image}, and further inspires us to investigate the ratio of $\mathbf{D^G}$ size to $\mathbf{D^S}$ size, denoted as $r$. 

\noindent \textbf{External versus Internal} Table III is one more set of convincing results to demonstrate the complementary behaviors of joint examples. Provided the total amount of dictionary atoms is fixed at 160, Table III lists how the average PSNR of SC LW  changes with $r$, where $\mathbf{D^G}$ has 160r/(1+r) atoms and $\mathbf{D^S}$ has 160/(1+r) atoms (previously $r$ = 4). As shown by Table III, increasing $r$ from 4 to 7 leads to improved PNSRs in heavy noise cases ($\sigma$ = 40 and 50). However, neither an overly large nor a small $r$ leads to any performance gain. On the one hand, the PSNR decreases rapidly with $r$ when $r \le$ 4, under all $\sigma$s. It proves the key role of external examples in reconstructing high-quality patches under mild noise conditions. On the other hand, the PSNR also becomes poor when $r$ = 15. The study of $r$ offers another powerful support for the importance of learning composite dictionaries from joint examples.

 \begin{table}[h]
 \begin{center}
 \caption{Comparison of PNSRs (dB) of SC LW under different $r$ and $\sigma$  values.}
 \vspace{1em}
 \begin{tabular}{|c|c|c|c|c|c|}
 \hline
 &  $\sigma$ = 10 &  $\sigma$ = 20 & $\sigma$ = 30  & $\sigma$ = 40 &  $\sigma$ = 50  \\
 \hline
r=0 & 30.47 &  28.48  & 27.67 & 26.20 & 24.06 \\
 \hline
r=1 & 32.23 &  30.27 & 29.95 &  26.82 & 25.03 \\
 \hline
r=3 & 35.46 & 33.27 & 31.80 &  28.97 & 26.21 \\
 \hline
r=4 & \textbf{36.27} & \textbf{34.24} & \textbf{32.83} &  28.76 &  26.32 \\
 \hline
r=7 & 36.18 & 34.05 & 32.58 &  \textbf{28.94} & \textbf{26.37} \\
 \hline
r=9 & 36.07 & 33.84 &  31.73 &  28.31 &  26.02 \\
 \hline
 r=15 & 34.57 & 31.28 &  30.80 &  27.67 &  25.83 \\
 \hline
 \end{tabular}
 \end{center}
 \end{table}
 

\subsection{Application II: Image SR}

The proposed method can be applied to solving image SR problems by a variant extension. First of all, the example pools are no longer collection of image patches, but instead \textit{example pairs} of a high-resolution (HR) patch and its low-resolution (LR) counterpart for each. Coupled dictionary learning has been proposed in \cite{Yang2012} to learn a \textit{dictionary pair} from a large external set of example pairs. To be formulated mathematically, the HR and LR patch spaces \{$\mathbf{X}_{ij}$\} and \{$\mathbf{Y}_{ij}$\} are assumed to be tied by some mapping function. With a well-trained coupled dictionary pair ($\mathbf{D_h}$, $\mathbf{D_l}$), it assumes that ($\mathbf{X}_{ij}$, $\mathbf{Y}_{ij}$) tends to admit a common sparse representation $\mathbf{a}_{ij}$. Yang et. al. \cite{Yang2012} suggested to first infer the sparse code $\mathbf{a}^L_{ij}$ of {$\mathbf{Y}_{ij}$} with respect to $\mathbf{D_l}$, and then use it as an approximation of $\mathbf{a}^H_{ij}$ (the sparse code of {$\mathbf{X}_{ij}$} with respect to $\mathbf{D_h}$), in order to recover $\mathbf{X}_{ij} \approx \mathbf{D_h} \mathbf{a}^L_{ij}$.

We construct the pool of external example pairs in the same way as \cite{Yang2012}. The pool of internal example pairs are less straightforward to construct, since the "groundtruth" HR image of the LR input is not available. To overcome the difficulty, we come up with an idea inspired by \cite{Fattal2010}. Based on the observation that singular features like edges and corners in small patches tend to repeat almost identically across different image scales, Freedman and Fattal \cite{Fattal2010} applied the ``high frequency transfer'' method to search the high-frequency component for a target HR patch, by NN patch matching across scales. Defining a linear interpolation operator $\mathcal{U}$ and a downsampling operator $\mathcal{D}$, for the input LR image $\mathbf{Y}$, we first obtain its initial upsampled image $\mathbf X^{'E} = \mathcal{U} (\mathbf{Y})$, and a smoothed input image $\mathbf{Y'} = \mathcal{D} (\mathcal{U} (\mathbf{Y}))$. Given the smoothed patch $\mathbf X_{ij}^{'E}$, the missing high-frequency band of each unknown patch $\mathbf X_{ij}^{E}$ is predicted by first solving a NN matching (\ref{freedman}):
\begin{equation}
\begin{array}{l}\label{freedman}
(m, n) = \arg\min_{(m, n) \in \mathcal{W}_{ij}} \| \mathbf Y'_{mn} - \mathbf X_{ij}^{'E}\|_F^2,
\end{array}
\end{equation}
where $\mathcal{W}_{ij}$ is defined as a small local searching window on image $\mathbf{Y'}$. With the co-located patch $\mathbf{Y}_{mn}$ from $\mathbf{Y}$, the high-frequency band $\mathbf{Y}_{mn} - \mathbf{Y'}_{mn}$ is pasted onto $\mathbf X_{ij}^{'E}$, i.e., $\mathbf X_{ij}^{E} = \mathbf X_{ij}^{'E} + \mathbf{Y}_{mn} - \mathbf{Y'}_{mn}$. Following this way, for the $(i,j)$-th patch of LR input $\mathbf{Y}$, we could treat $\mathbf X_{ij}^{E}$ as its corresponding HR patch and make them an internal example pair.  

  \begin{table}[h]
 \begin{center}
 \caption{Comparison of PNSRs (dB) used different SR methods.}
 \vspace{1em}
 \begin{tabular}{|c|c|c|c|}
 \hline
 &  \textit{Temple} &  \textit{Train} & \textit{Leopard} \\
 \hline
Bicubic & 25.29 &  26.14  & 24.14 \\
 \hline
Yang et.al.\cite{Yang2012} & 26.20 &  26.58 & 25.32  \\
 \hline
Freedman and Fattal \cite{Fattal2010} & 21.17 & 22.54 & 23.04 \\
 \hline
Proposed & \textbf{26.86} & \textbf{27.44} & \textbf{25.62} \\
 \hline
 \end{tabular}
 \end{center}
 \end{table}

We then apply the coupled dictionary learning algorithm in the similar manner (with the same $K$, $M$ and $N$) as using K-SVD above, obtaining the \textit{global dictionary pair} ($\mathbf{D^G_h}$, $\mathbf{D^G_l}$) and the \textit{sample-specific dictionary pair} ($\mathbf{D^S_h}$, $\mathbf{D^S_l}$). We solve (\ref{e00}) over $\mathbf{D^G_l}$ and $\mathbf{D^S_l}$, after which we export the sparse codes as well as the learned weight values, to reconstruct the final HR patches by $\mathbf{D^G_h}$ and $\mathbf{D^S_h}$. Comparison experiments are conducted against bicubic interpolation, Yang et.al.'s external example-based SR method \cite{Yang2012}, and Freedman et. al.'s internal example-based method \cite{Fattal2010}, on three test images \textit{Temple}, \textit{Train} and \textit{Leopard}, with a factor of 3. While our method is not specifically optimized for image SR, it obtains better SR results than the other two competitive methods \cite{Yang2012, Fattal2010}.

\section{Conclusion}
We propose a novel composite dictionary design framework. The composite dictionary consists of global and sample-specific parts, learned from external and internal examples, respectively. We formulate the similarity weights that adaptively correlate sparse codes with base dictionary atoms. Experiments demonstrate that the joint utilization of external and internal examples outperforms either stand-alone alternative. The applications in image denoising and SR show great potential along this research line.


\bibliographystyle{IEEEbib}
\bibliography{refs}

\begin{thebibliography}{10}

\bibitem{elad2006image}
Michael Elad and Michal Aharon,
\newblock ``Image denoising via sparse and redundant representations over
  learned dictionaries,''
\newblock {\em Image Processing, IEEE Transactions on}, vol. 15, no. 12, pp.
  3736--3745, 2006.

\bibitem{rubinstein2010dictionaries}
Ron Rubinstein, Alfred~M Bruckstein, and Michael Elad,
\newblock ``Dictionaries for sparse representation modeling,''
\newblock {\em Proceedings of the IEEE}, vol. 98, no. 6, pp. 1045--1057, 2010.

\bibitem{yang2010image}
Jianchao Yang, John Wright, Thomas~S Huang, and Yi~Ma,
\newblock ``Image super-resolution via sparse representation,''
\newblock {\em Image Processing, IEEE Transactions on}, vol. 19, no. 11, pp.
  2861--2873, 2010.

\bibitem{Fattal2010}
Gilad Freedman and Raanan Fattal,
\newblock ``Image and video upscaling from local self-examples,''
\newblock {\em ACM Transactions on Graphics (TOG)}, vol. 30, no. 2, pp. 12,
  2011.

\bibitem{wang2014joint}
Zhangyang Wang, Zhaowen Wang, Shiyu Chang, Jianchao Yang, and Thomas Huang,
\newblock ``A joint perspective towards image super-resolution: Unifying
  external-and self-examples,''
\newblock in {\em Applications of Computer Vision (WACV), Winter Conference
  on}. IEEE, 2014, pp. 596--603.

\bibitem{wang2015learning}
Zhangyang Wang, Yingzhen Yang, Zhaowen Wang, Shiyu Chang, Jianchao Yang, and
  T.S. Huang,
\newblock ``Learning super-resolution jointly from external and internal
  examples,''
\newblock {\em Image Processing, IEEE Transactions on}, vol. 24, no. 11, pp.
  4359--4371, Nov 2015.

\bibitem{mairal2012task}
Julien Mairal, Francis Bach, and Jean Ponce,
\newblock ``Task-driven dictionary learning,''
\newblock {\em Pattern Analysis and Machine Intelligence, IEEE Transactions
  on}, vol. 34, no. 4, pp. 791--804, 2012.

\bibitem{mairal2008discriminative}
Julien Mairal, Francis Bach, Jean Ponce, Guillermo Sapiro, and Andrew
  Zisserman,
\newblock ``Discriminative learned dictionaries for local image analysis,''
\newblock in {\em Computer Vision and Pattern Recognition, 2008. CVPR 2008.
  IEEE Conference on}. IEEE, 2008, pp. 1--8.

\bibitem{yang2014latent}
Meng Yang, Dengxin Dai, Lilin Shen, and Luc~Van Gool,
\newblock ``Latent dictionary learning for sparse representation based
  classification,''
\newblock in {\em Computer Vision and Pattern Recognition (CVPR), 2014 IEEE
  Conference on}. IEEE, 2014, pp. 4138--4145.

\bibitem{yang2011exploiting}
Chih-Yuan Yang, Jia-Bin Huang, and Ming-Hsuan Yang,
\newblock ``Exploiting self-similarities for single frame super-resolution,''
\newblock in {\em Computer Vision--ACCV 2010}, pp. 497--510. Springer, 2011.

\bibitem{chatterjee2010denoising}
Priyam Chatterjee and Peyman Milanfar,
\newblock ``Is denoising dead?,''
\newblock {\em Image Processing, IEEE Transactions on}, vol. 19, no. 4, pp.
  895--911, 2010.

\bibitem{zontak2011internal}
Maria Zontak and Michal Irani,
\newblock ``Internal statistics of a single natural image,''
\newblock in {\em Computer Vision and Pattern Recognition (CVPR), 2011 IEEE
  Conference on}. IEEE, 2011, pp. 977--984.

\bibitem{mosseri2013combining}
Inbar Mosseri, Maria Zontak, and Michal Irani,
\newblock ``Combining the power of internal and external denoising,''
\newblock in {\em Computational Photography (ICCP), International Conference
  on}. IEEE, 2013, pp. 1--9.

\bibitem{wang2015self}
Zhangyang Wang, Yingzhen Yang, Zhaowen Wang, Shiyu Chang, Wei Han, Jianchao
  Yang, and Thomas Huang,
\newblock ``Self-tuned deep super resolution,''
\newblock in {\em Proceedings of the IEEE Conference on Computer Vision and
  Pattern Recognition Workshops}, 2015, pp. 1--8.

\bibitem{rubinstein2010double}
Ron Rubinstein, Michael Zibulevsky, and Michael Elad,
\newblock ``Double sparsity: Learning sparse dictionaries for sparse signal
  approximation,''
\newblock {\em Signal Processing, IEEE Transactions on}, vol. 58, pp.
  1553--1564, 2010.

\bibitem{lee2006efficient}
Honglak Lee, Alexis Battle, Rajat Raina, and Andrew~Y Ng,
\newblock ``Efficient sparse coding algorithms,''
\newblock in {\em Advances in neural information processing systems}, 2006, pp.
  801--808.

\bibitem{kamada2006support}
Yuya Kamada and Shigeo Abe,
\newblock ``Support vector regression using mahalanobis kernels,''
\newblock in {\em Artificial Neural Networks in Pattern Recognition}, pp.
  144--152. Springer, 2006.

\bibitem{Yang2012}
Jianchao Yang, Zhaowen Wang, Zhe Lin, Scott Cohen, and Thomas Huang,
\newblock ``Coupled dictionary training for image super-resolution,''
\newblock {\em Image Processing, IEEE Transactions on}, vol. 21, no. 8, pp.
  3467--3478, 2012.

\end{thebibliography}

\end{document}